\documentclass[journal]{IEEEtran}
\usepackage{amsmath,amsfonts}
\usepackage{algorithmic}
\usepackage{algorithm}
\usepackage{array}
\usepackage[caption=false,font=normalsize,labelfont=sf,textfont=sf]{subfig}
\usepackage{textcomp}
\usepackage{stfloats}
\usepackage{url}
\usepackage{verbatim}
\usepackage{graphicx}
\usepackage{cite}
\usepackage{xcolor}
\usepackage{tabu}                     
\usepackage{multirow}                 
\usepackage{multicol}                 
\usepackage{multirow}                
\usepackage{float}                    
\usepackage{makecell}                 
\usepackage{booktabs}                 
\usepackage{subfig}
\usepackage{bbding}

\usepackage{soul}
\usepackage{stfloats}
\usepackage{footnote}
\usepackage{xr} 

\begin{document}
	
	\title{

		Expose Camouflage in the Water: Underwater Camouflaged Instance Segmentation and  Dataset 

}
	
	\author{Chuhong Wang, 
		Hua Li,~\IEEEmembership{Member,~IEEE,} 
		Chongyi Li,~\IEEEmembership{Senior Member,~IEEE,} 
        Huazhong Liu,~\IEEEmembership{Member,~IEEE,} 
        Xiongxin Tang,
		and Sam Kwong,~\IEEEmembership{Fellow,~IEEE}

	\IEEEcompsocitemizethanks{
		\IEEEcompsocthanksitem Chuhong Wang is with the School of Information and Communication Engineering, Hainan University, Haikou 570228, China, and also with the School of Electronic and Information Engineering, Guangdong Ocean University, Zhanjiang 524088, China (e-mail: wangchuhong@hainanu.edu.cn).
		\IEEEcompsocthanksitem Hua Li and Huazhong Liu are with the School of Computer Science and Technology, Hainan University, Haikou 570228, China (e-mail: lihua@hainanu.edu.cn, hzliu@hainanu.edu.cn).
		\IEEEcompsocthanksitem Chongyi Li is with the School of Computer Science, Nankai University, Tianjin 300071, China (e-mail: lichongyi25@gmail.com).     
        \IEEEcompsocthanksitem Xiongxin Tang is with the Institute of Software, Chinese Academy of Science, Beijing 100190, China (e-mail: xiongxin@iscas.ac.cn).
		\IEEEcompsocthanksitem Sam Kwong is with the School of Data Science, Lingnan University, Hong Kong, SAR, China (e-mail: samkwong@ln.edu.hk).
	
	}
}

	\markboth{Journal of \LaTeX\ Class Files,~Vol.~14, No.~8, August~2021}%
	{Shell \MakeLowercase{\textit{et al.}}: A Sample Article Using IEEEtran.cls for IEEE Journals}
	
	
	\maketitle
	
	\begin{abstract}
	 
	 With the development of underwater exploration and marine protection, underwater vision tasks are widespread. 
     Due to the degraded underwater environment, characterized by color distortion, low contrast, and blurring, camouflaged instance segmentation (CIS) faces greater challenges in accurately segmenting objects that blend closely with their surroundings.
	 Traditional camouflaged instance segmentation methods, trained on terrestrial-dominated datasets with limited underwater samples, may exhibit inadequate performance in underwater scenes.
	 To address these issues, we introduce the first underwater camouflaged instance segmentation (UCIS) dataset, abbreviated as UCIS4K, which comprises 3,953 images of camouflaged marine organisms with instance-level annotations.
     In addition, we propose an Underwater Camouflaged Instance Segmentation network based on Segment Anything Model (UCIS-SAM). 
     Our UCIS-SAM includes three key modules.
	 First, the Channel Balance Optimization Module (CBOM) enhances channel characteristics to improve underwater feature learning, effectively addressing the model's limited understanding of underwater environments.
	 Second, the Frequency Domain True Integration Module (FDTIM) is proposed to emphasize intrinsic object features and reduce interference from camouflage patterns, enhancing the segmentation performance of camouflaged objects blending with their surroundings. 
     Finally, the Multi-scale Feature Frequency Aggregation Module (MFFAM) is designed to strengthen the boundaries of low-contrast camouflaged instances across multiple frequency bands, improving the model's ability to achieve more precise segmentation of camouflaged objects.
	 Extensive experiments on the proposed UCIS4K and public benchmarks show that our UCIS-SAM outperforms state-of-the-art approaches. 
	 The code and dataset are released at https://github.com/wchchw/UCIS4K.
	\end{abstract}	
	
	\section{Introduction}

	\IEEEPARstart{C}{amouflage} is a biological strategy whereby an organism alters its physical appearance to blend in with its surroundings, thereby reducing visibility and increasing the likelihood of avoiding detection or predation \cite{zhou2022feature}. Camouflaged instance segmentation (CIS) aims to accurately identify and segment camouflaged instances from surroundings.	
	These instances skillfully employ color, texture, and shape to minimize contrast with the background, rendering feature extraction highly complex and challenging \cite{he2024text,fu2024semi}. 
	The edges of camouflaged instances blend almost seamlessly with the background, lacking clear boundaries, which significantly increases the difficulty of instance segmentation \cite{yin2024camoformer,zhang2023predictive}.    
	With the rapid advancements in deep learning for visual technologies \cite{yang2025shell,liao2024image,jin2020deep}, the increasing demand for underwater exploration has driven the development of Underwater Camouflaged Instance Segmentation (UCIS).
	The primary goal of UCIS is to improve segmentation accuracy and analytical capabilities in underwater environments, with applications including ecological preservation, and underwater exploration.	
	\begin{figure}[!t]
		\centering	
		\vspace{-0.15cm} 
		\captionsetup[subfigure]{font=scriptsize}
		\subfloat[Proportion]{			
		\raisebox{0.098\height}
		{\includegraphics[width=0.175\textwidth]{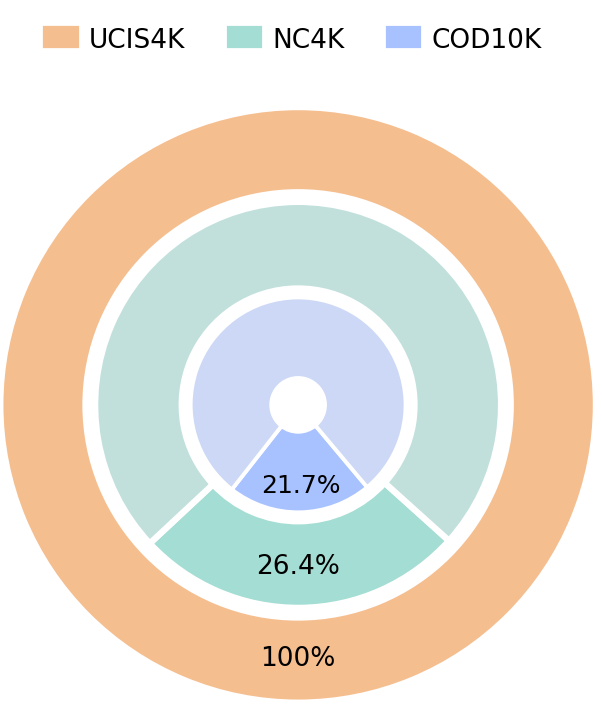}%
			\label{example_1}}}
		\hfil
		\subfloat[Result comparison]
		{\includegraphics[width=0.3\textwidth]{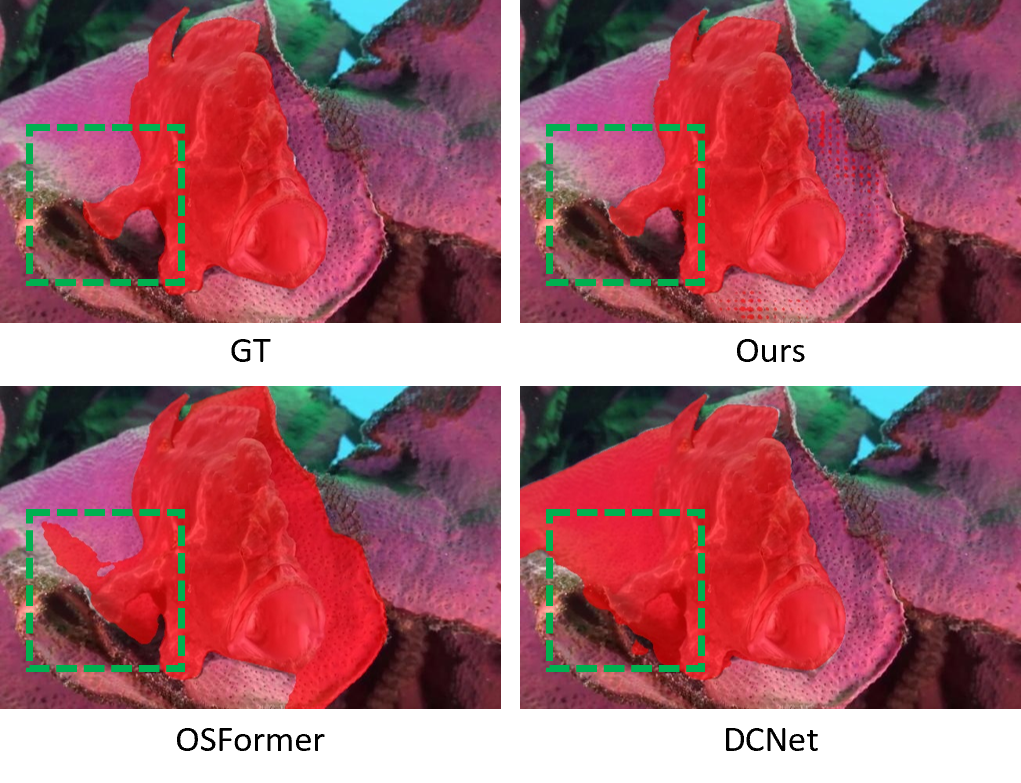}%
			\label{example_2}}
		
		\caption{A comparative analysis of our dataset and method against existing datasets and methods.  (a) The proportion of underwater images in the our UCIS4K, COD10K\cite{fan2020camouflaged}, and NC4K\cite{lv2021simultaneously}. (b) Comparison of segmentation results. The CIS models OSFormer \cite{pei2022osformer} and DCNet \cite{luo2023camouflaged} confuse the instance with underwater surroundings, while ours can segment it more accurately.}
		\label{contrast1}
	\end{figure}

	However, UCIS faces challenges due to the limited availability of specialized underwater camouflage datasets, which are essential for effective model training. 
	As illustrated in Fig. \ref{contrast1}\subref{example_1}, existing camouflaged instance segmentation datasets COD10K\cite{fan2020camouflaged} and NC4K\cite{lv2021simultaneously} contain only a limited number of underwater images and are not specifically designed for underwater environments.
	Consequently, the models developed for these general CIS datasets tend to show a performance decline in underwater scenarios.
	As a case shown in Fig.~\ref{contrast1}\subref{example_2}, the performance of state-of-the-art CIS methods OSFormer \cite{pei2022osformer} and DCNet \cite{luo2023camouflaged} is degraded. These models fail to effectively distinguish between underwater backgrounds and the object. 
     The lack of such datasets notably restricts the development and fine-tuning of models for precise underwater instance segmentation, hindering progress in the field of UCIS.

	Furthermore, underwater images are affected by the distinct properties of the transmission medium, which presents challenges in image processing \cite{cong2023pugan,chen2024underwater}.	
	Increased depth of water causes illumination decay, resulting in uneven brightness and a shift towards blue-green hues \cite{rao2023deep}. Backscatter reduces contrast, while forward scattering blurs edges \cite{zhuang2022underwater}. 
	Moreover, water currents, plankton, and suspended particles introduce noise, further degrading image clarity\cite{chen2024cwscnet}. These challenges complicate the development of camouflaged instance segmentation models for underwater environments.	
	The general underwater instance segmentation model, WaterMask \cite{lian2023watermask}, although not specifically designed for camouflaged instances, demonstrates relatively better performance in distinguishing objects from the surrounding underwater environment. However, it still faces challenges in accurately capturing fine details of camouflaged instances, especially in regions where textures and colors closely resemble the background, as well as fuzzy boundary issues. These limitations lead to insufficient segmentation accuracy, ultimately restricting its effectiveness for tasks involving camouflaged instances.

	\begin{figure*}[!t]
		\centering
		\vspace{-0.25cm} 
		\includegraphics[width=0.98\textwidth]{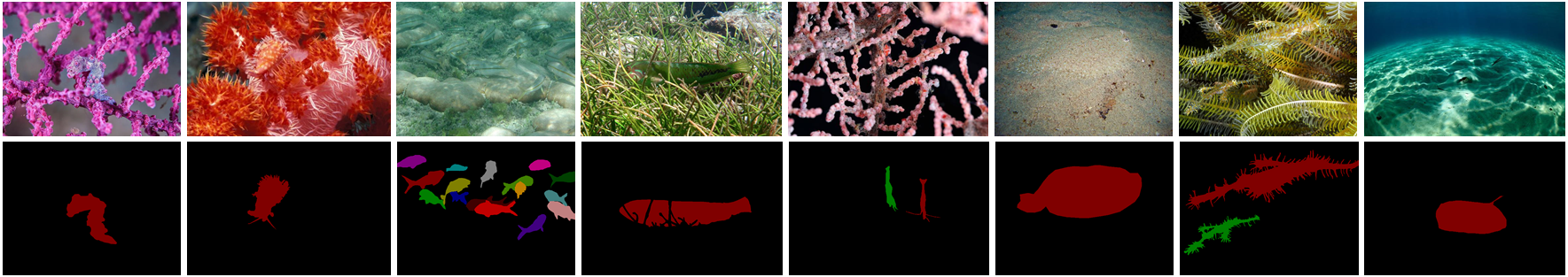}
		\caption{Examples of various challenging attributes from our UCIS4K dataset.
		It includes camouflaged objects with similar colors and textures to the background, blurred contours, small sizes, multiple objects, occlusion, complex contours, transparency, and underwater scenes with light and shadow effects.}
		\vspace{-5pt}

		\label{dataset}
	\end{figure*}

	To alleviate the aforementioned issues, we construct the first Underwater Camouflaged Instance Segmentation dateset UCIS4K, aiming at stimulating the exploration of camouflaged instance segmentation in underwater scenes. 
	The UCIS4K dataset consists of 3,953 camouflaged images, encompassing a diverse array of marine organisms, such as fish, shrimp, crabs, and seahorses, across various camouflaged scenarios. 
	As illustrated in Fig. \ref{dataset}, the dataset employs diverse camouflage mechanisms annotated with instance-level masks, including background-matching colors and textures, indistinct contours, diminutive object sizes, multiple objects, occlusion, intricate shapes, and shadow effects in underwater environments.	

	Moreover, we propose an underwater camouflaged instance segmentation architecture based on the Segment Anything Model (UCIS-SAM). Most existing methods for camouflaged instance segmentation rely on spatial-domain processing, such as multi-scale feature fusion\cite{pei2022osformer}, contour-focused feature extraction\cite{nguyen2023ost}, and attention mechanisms\cite{dong2023unified}. 
	Although these approaches have enhanced the model's ability to perceive camouflaged objects, they still face limitations in fully capturing the confusing details in underwater scenes.
	The Segment Anything Model (SAM)\cite{kirillov2023segment}, which achieves remarkable performance in image segmentation through large-scale pre-training and multi-modal prompting, shows the potential to address the above limitations.
	Nevertheless, its performance may be limited in specific domains due to the absence of domain-specific knowledge \cite{chen2023sam}.
	To address the color distortion in underwater environments, we integrate the Channel Balance Optimization Module (CBOM) into SAM’s encoder to adjust feature learning, compensating for the model's lack of underwater environmental knowledge and enhancing its performance in underwater scenarios.
	Then, we propose a frequency-domain-based approach to tackle the challenge of high similarity in texture and color between objects and background in underwater camouflaged scenarios.
	Specifically, we introduce the Frequency Domain True Integration Module (FDTIM) to improve the model's ability to segment camouflaged instances by maximizing the intrinsic features of the object and
	reducing affect from the similar surrounding environments. This approach effectively overcomes the limitations of traditional spatial-domain methods.
	Moreover, we devise the Multi-scale Feature Frequency Aggregation Module (MFFAM), which sharpens the boundaries of low-contrast camouflaged instances by analyzing fine-scale details through high-frequency features. Meanwhile, low-frequency features capture the overall structure and generate salient prompts to guide SAM's mask decoder.
	
	Extensive experiments are conducted to validate the effectiveness of our UCIS-SAM model and the proposed UCIS4K dataset. First, we compared UCIS-SAM with the state-of-the-art method on UCIS4K dataset. Then, we perform the comparison experiments on CIS datasets COD10K\cite{fan2020camouflaged} and NC4K\cite{lv2021simultaneously}, and the underwater instance dataset segmentation UIIS\cite{lian2023watermask} to verify the generalization ability. 
	The main contributions are concluded as follows:

	\begin{itemize}
		\item We contribute the first dataset UCIS4K for the underwater camouflaged instance segmentation task, which encompasses 3,953 images with instance-level annotations. It captures the diverse appearances of camouflaged organisms in underwater environments, highlighting the characteristics of camouflage in underwater scenes.
		\item We propose UCIS-SAM for underwater CIS task, incorporating CBOM into SAM's encoder to mitigate color distortion and adjust feature learning, thereby achieving effective domain adaptation to underwater environments.
		\item We propose FDTIM to alleviate the affect from high similarity with the surrounding environment, and MFFAM to enhance the boundaries of low-contrast camouflaged instances, enabling the model to acquire camouflage-specific knowledge and improve segmentation accuracy.
		\item Comprehensive experiments on public benchmarks and datasets have verified the effectiveness of the proposed UCIS-SAM model and UCIS4K dataset.

	\end{itemize}

	\section{Related Work}\label{section2}

    \subsection{Camouflaged Instance Segmentation}	
	
		\begin{figure}[!t]
		\centering	
		\vspace{-0.25cm} 
		\captionsetup[subfigure]{font=scriptsize}
		\subfloat[]{\includegraphics[width=0.118\textwidth]{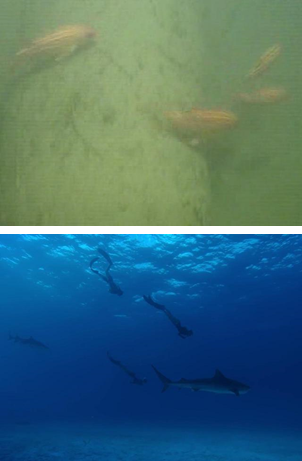}%
			\label{example_aa}}
		\hfil
		\subfloat[]{\includegraphics[width=0.118\textwidth]{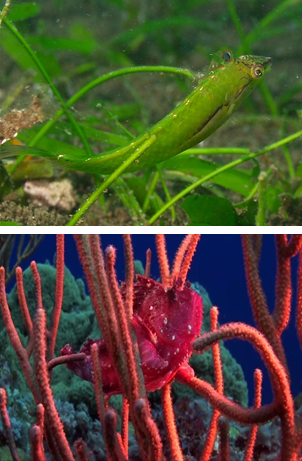}%
			\label{example_bb}}
		\hfil
		\subfloat[]{\includegraphics[width=0.118\textwidth]{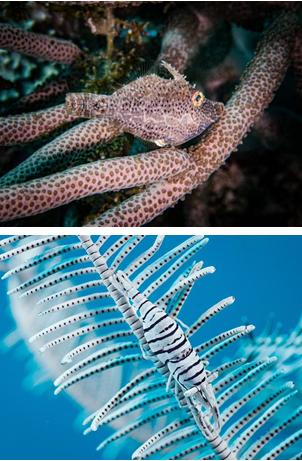}%
			\label{example_cc}}
		\hfil
		\subfloat[]{\includegraphics[width=0.118\textwidth]{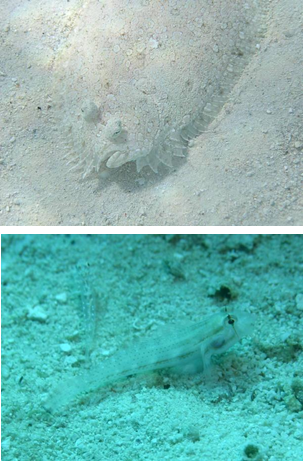}%
			\label{example_dd}}
		\caption{Examples of uncamouflaged objects and camouflaged objects. (a) Uncamouflaged objects appear unclear due to motion or backlighting. (b) Color camouflaged objects, (c) Texture camouflaged objects, (d) Edge blur camouflaged objects.}
		\label{example111}
	\end{figure}
	\begin{figure*}[!b]
		\centering
		\vspace{-0.45cm} 
		\captionsetup[subfigure]{font=scriptsize}
		\subfloat[]{\includegraphics[width=0.33\textwidth]{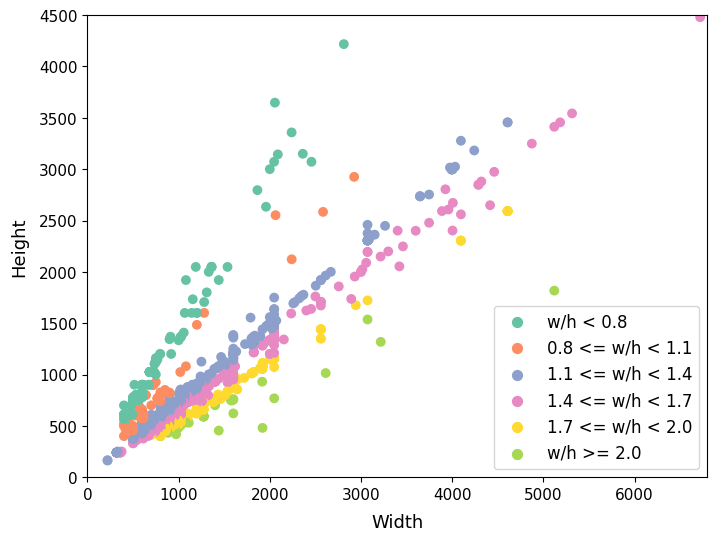}%
			\label{resolution_ucis}}
		\hfil
		\subfloat[]{\includegraphics[width=0.33\textwidth]{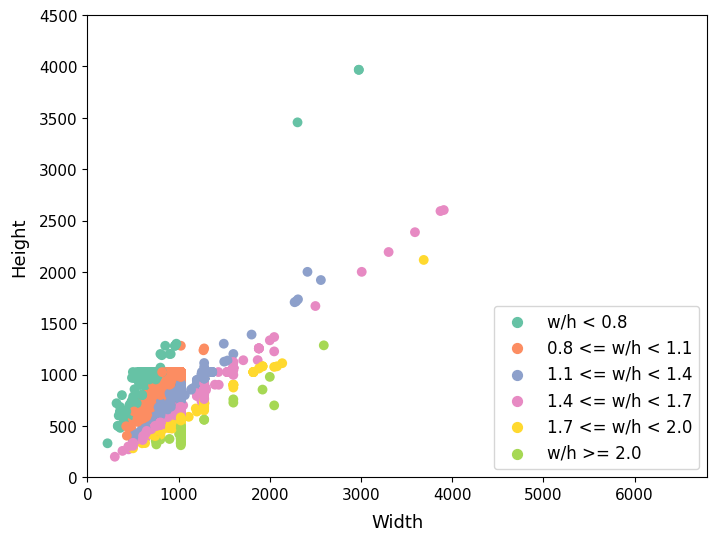}%
			\label{resolution_cod10k}}
		\hfil
		\subfloat[]{\includegraphics[width=0.33\textwidth]{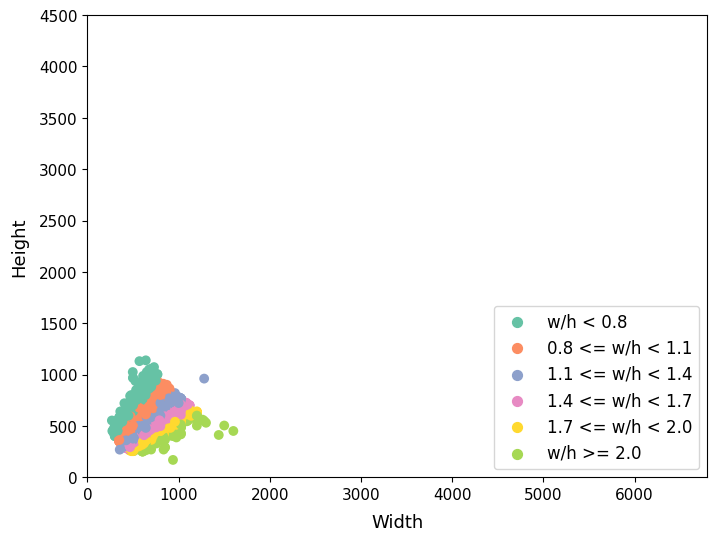}%
			\label{resolution_nc4k}}
		\vspace{-0.1cm} 
		\caption{The resolution distribution of images in the  camouflaged dataset. (a) UCIS4K, (b) COD10K\cite{fan2020camouflaged}, (c) NC4K\cite{lv2021simultaneously}.
			Our UCIS4K dataset contains higher-resolution images than both COD10K and NC4K, providing richer visual information.}
		\label{resolution}
		
	\end{figure*}
	Camouflaged instance segmentation (CIS) involves accurately identifying and segmenting instances in highly complex and variable natural environments. Although current research has made certain advancements, existing CIS datasets and methods primarily focus on terrestrial scenes.	
	The CAMO dataset \cite{le2019anabranch} is the first camouflage dataset with more than 1,000 annotated images, followed by instance-level annotation\cite{le2021camoufinder}.
	It is then extended to CAMO++ \cite{le2021camouflaged} for CIS task, which contains 2,700 camouflaged images. Meanwhile, a simple yet effective camouflage fusion learning framework was proposed by leearning image context.	
	The COD10K dataset \cite{fan2020camouflaged} is a milestone in the field, providing 3,040 high-quality instance-level camouflaged training images and 2,026 testing images.
	Furthermore, the NC4K\cite{lv2021simultaneously} dataset provides 4,121 camouflaged images for testing.
	Currently, the majority of CIS networks are trained and evaluated on the two benchmark datasets, COD10K and NC4K, which primarily focus on terrestrial organisms.
	OSFormer \cite{pei2022osformer} introduces a location-sensing transformer to seize instance clues at different locations and a coarse-to-fine fusion module to integrate multi-scale features, enabling one-stage camouflaged instance segmentation.
	CE-OST \cite{nguyen2023ost} employs transformer-based models to boost the performance by enhancing the contours of camouflaged instances.
	UQFormer \cite{dong2023unified} innovates a unified query-based paradigm for CIS, integrating global camouflaged object region and boundary cues in a multi-task learning framework.		
	DCNet \cite{luo2023camouflaged} introduces a pixel-level camouflage decoupling module that utilizes a differential attention mechanism to mitigate the characteristics of camouflage, alongside an instance-level camouflage suppression module which integrates reliable reference points to construct a more robust similarity metric.		
	GLNet \cite{li2024camouflaged} features a dual-branch convolutional feed-forward network for global capture and edge-guide fusion modules for local refinement to discern camouflaged instance details.	
	TPNet \cite{he2024text} is a weakly-supervised camouflaged instance segmentation method that leverages text prompts and semantic distinctions, comprising pseudo mask generation and self-training stages for effective segmentation.	
	AQSFormer \cite{dong2024adaptive} is proposed to address query redundancy by selecting valid queries adaptively and incorporating boundary positional embedding for improved accuracy.

	\subsection{Segment Anything Model and Its Applications}
	SAM \cite{kirillov2023segment}, developed by Meta AI, is a foundational segmentation model trained on over one billion annotations, enabling zero-shot generalization to new tasks through prompt engineering. Its strong performance and high segmentation accuracy in natural image segmentation have made it widely adopted across various fields\cite{zhang2024efficientvit}.
	However, SAM's performance is limited in certain domains, requiring adaptations in domain-specific applications to meet their unique tasks and contextual requirements\cite{xu2024eviprompt}.
	In medical imaging,
	the H-SAM \cite{cheng2024unleashing} leverages a two-stage decoder with mask-guided self-attention, learnable mask cross-attention, and a hierarchical pixel decoder to improve segmentation accuracy and detail.
	The MA-SAM \cite{chen2024ma} injects a series of 3D adapters into the transformer blocks, enabling the pre-trained 2D backbone to extract 3D information from input data.
	While in remote sensing, researchers optimize input prompts and develop methods to enhance SAM's task-specific performance\cite{chen2024rsprompter}.
	An auxiliary optimization strategy\cite{ma2024sam} for SAM is developed to enhance semantic segmentation performance by introducing object consistency and boundary preservation losses.
	Within the agricultural domain,  
	researchers have introduced a methodology for crop segmentation based on SAM, employing a multistage adaptive fine-tuning process to enhance its performance on agricultural imagery\cite{song2024multispectral}.
	Similarly, the complex lighting conditions and noise interference characteristic of underwater environments, coupled with the low contrast and blurry boundaries associated with camouflage, pose substantial challenges to the segmentation performance of SAM.

	\begin{figure}[!t]
	\centering
	\vspace{-0.35cm} 
	\includegraphics[width=0.38\textwidth]{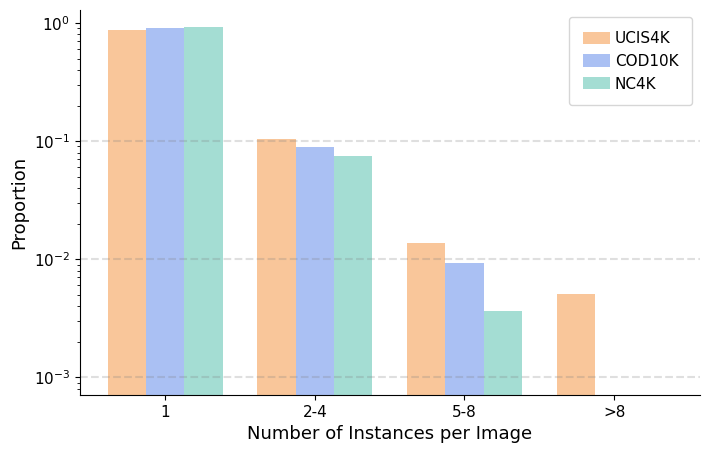}
	\vspace{-0.35cm} 
	\caption{The distribution of the number of camouflaged instances per image in the UCIS4K, COD10K\cite{fan2020camouflaged}, and NC4K\cite{lv2021simultaneously} dataset.}
	\label{number}
\end{figure}

\begin{figure}[!t]
	\setlength{\abovecaptionskip}{-2pt}
	\vspace{-0.35cm} 
	\centering
	\includegraphics[width=0.36\textwidth]{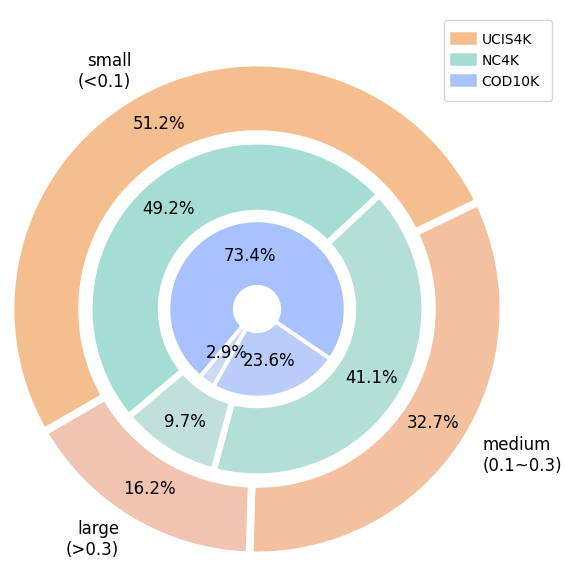}
	\vspace{-0.15cm} 
	\caption{The mask size distribution of camouflaged instances in the UCIS4K, COD10K\cite{fan2020camouflaged}, and NC4K\cite{lv2021simultaneously} dataset.}
	\label{size}
	\vspace{-0.15cm} 
\end{figure}

	\section{UCIS4K Dataset}\label{section3}
	\subsection{Dataset Collection and Annotation}
	To construct an underwater camouflage image dataset, we initially collected approximately 9,000 images of underwater organisms from the public underwater datasets and images using camouflage-related keywords. A total of 3,953 images were selected by trained volunteers based on camouflage characteristics.
	These images were then annotated at the pixel level, with the results validated through a voting process among the volunteers. Overall, instance-level annotations were successfully completed on 3,953 images for the UCIS4K dataset. As shown in Fig. \ref{dataset}, the dataset encompasses a wide range of complex scenarios, providing a comprehensive resource for training and	evaluating models designed to segment camouflaged objects under varied conditions.
	In this context, a camouflaged object is defined as one whose color, texture, or structure blends with the surrounding environment (Fig. \ref{example111}\subref{example_bb} and \subref{example_cc}), or whose edges are blurred (Fig. \ref{example111}\subref{example_dd}), making it difficult to distinguish from the background. 
	In contrast, Fig. \ref{example111}\subref{example_aa} are uncamouflaged objects, which appear unclear due to motion or backlighting.

	\begin{figure}[!t]
		\centering
		\vspace{-0.35cm} 
		\includegraphics[width=0.42\textwidth]{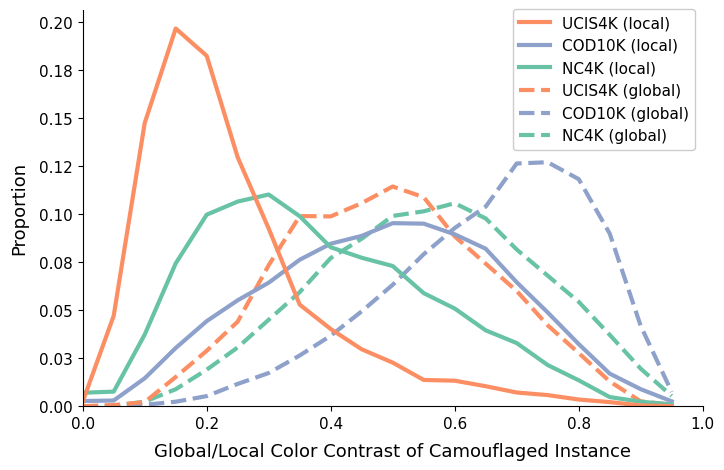}
		\vspace{-0.35cm} 
		\caption{The comparison of UCIS4K, COD10K\cite{fan2020camouflaged}, and NC4K\cite{lv2021simultaneously} in global color contrast and local color contrast.}
		\label{contrast}
	\end{figure}

	\subsection{Dataset Features and Statistics}	
	\subsubsection{Image Resolution}
	The image resolution of the UCIS4K dataset spans a wide range, from $220 \times 162$ pixels to $6720 \times 4480$ pixels. As shown in Fig. \ref{resolution}, the UCIS4K dataset contains more high-resolution images than both the COD10K\cite{fan2020camouflaged} and NC4K\cite{lv2021simultaneously} datasets. 
	This attribute provides a notable advantage by offering a richer array of visual information and more nuanced image features, as high-resolution images capture a greater level of detail, enhancing model training.
	\subsubsection{The Number of Camouflaged Instances}
	In the UCIS4K dataset, each image contains one to multiple instances of camouflage, with some images featuring over forty instances. 
	As shown in Fig. \ref{number}, the proportion of images with 5 to 8 instances exceeds 1\%, and those with 2 to 4 instances surpass 10\%, both of which are higher than the corresponding ratios in the COD10K\cite{fan2020camouflaged} and NC4K\cite{lv2021simultaneously} datasets. 
	It is worth noting that approximately 0.5\% of the images in the UCIS4K dataset contain more than 8 instances, which is absent in the other two datasets.
	This also means that the UCIS4K dataset presents a greater challenge for camouflaged instance segmentation, especially in handling high-density instances.

	\subsubsection{The Mask Size of Camouflaged Instance}
	The mask size of an instance is defined by the proportion of pixels constituting the mask relative to the total pixel count of the  image\cite{le2021camouflaged}.
	Our UCIS4K dataset covers a wide range of scales, from 0.007\% to 93.787\%. 
	As presented in Fig. \ref{size}, small instances (less than 0.1) account for 51.2\%, and medium instances (ranging from 0.1 to 0.3) make up 32.7\%. This distribution pattern is consistent with that of existing camouflage datasets, such as COD10K\cite{fan2020camouflaged} and NC4K\cite{lv2021simultaneously}, which also exhibit a size distribution where small and medium instances are more abundant, while large instances are relatively scarce.

	\subsubsection{The Degree of Camouflage in Instances}
	Considering the effectiveness of camouflage, we have identified the contrast between an object and its background as a key factor, where lower contrast indicates stronger camouflage.
	The global contrast of the RGB histograms for both the camouflaged object and its background \cite{lian2024diving,fan2018salient} is calculated to measure the difference between them using the Bhattacharyya distance \cite{choi2003feature}. As shown in Fig. \ref{contrast}, the UCIS4K dataset exhibits a lower global contrast relative to the background, indicating a more pronounced camouflage effect compared to the COD10K\cite{fan2020camouflaged} and NC4K\cite{lv2021simultaneously} datasets.	
	Furthermore, a significant challenge in camouflaged instance segmentation lies in delineating object boundaries, as the similarity between the camouflaged object and its surrounding environment makes the boundary areas difficult to distinguish. 
	By calculating the local contrast of a $5\times5$ patch at the boundary of each camouflaged object \cite{li2014secrets}, we find that camouflaged objects in the UCIS4K dataset are more effectively concealed, thus imposing higher demands on the accuracy of camouflaged instance segmentation.

    More details about the UCIS4K dataset are provided in the supplementary materials.

\begin{figure*}[!t]
	\centering
	\vspace{-0.25cm} 
	\includegraphics[width=0.98\textwidth]{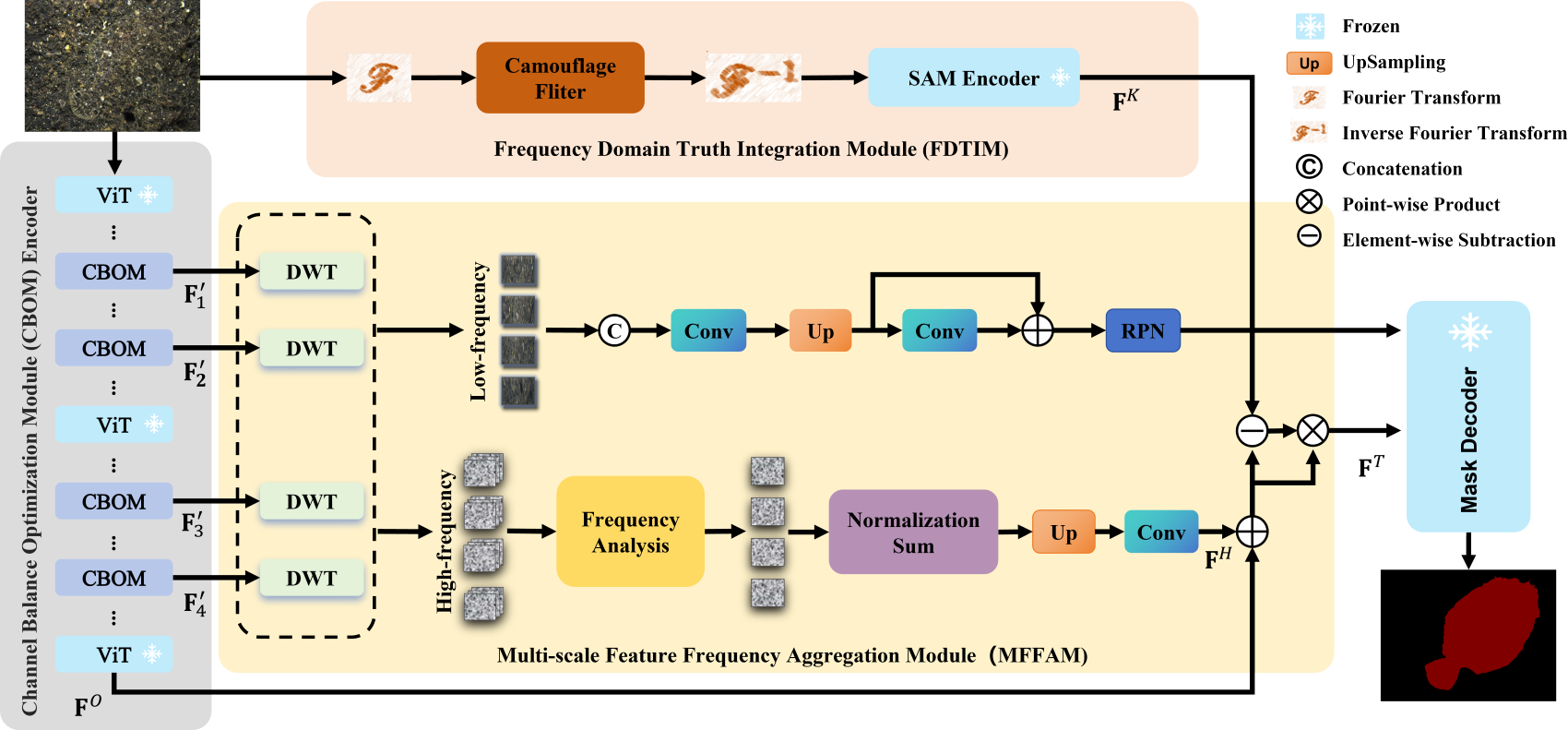}
	\vspace{-0.15cm} 
	\caption{The overall framework of UCIS-SAM consists of three main components: The CBOM encoder integrates the CBOM to adjust underwater feature learning; the FDTIM reduces the interference of camouflage patterns in the frequency domain to learn camouflage-specific domain knowledge; the MFFAM aggregates multi-level features to generate salient prompts and enhance boundary details for more accurate segmentation.}

	\label{framework}
\end{figure*}

	\section{The Proposed UCIS-SAM}\label{section4}
	\subsection{Overall Architecture}
	The overall framework of our UCIS-SAM model is illustrated in Fig. \ref{framework}. 
	Given an input underwater image, it is first processed by an encoder integrated with the CBOM, which aims to correct chromatic discrepancies caused by underwater conditions such as water turbidity and light attenuation. By adjusting color accuracy and modulating image channel properties, the CBOM encoder generates a feature map $\mathbf{F}^O$ with more reliable and balanced color information, improving segmentation of underwater objects.
	Simultaneously, the input image is also passed through the FDTIM, which isolates camouflaged features by filtering background noise while preserving relevant non-camouflaged information. The resulting feature map $\mathbf{F}^{K}$ enhances the extraction of the object's intrinsic features and mitigates interference from camouflage patterns that closely resemble the surrounding environment.	
	
	The feature map $\mathbf{F}'$ from the CBOM is then fed into the MFFAM, which aggregates multi-level features derived from both low-frequency and high-frequency components using Discrete Wavelet Transform (DWT).
	The low-frequency components $\mathbf{F}^L$ provide global contextual information, generating salient prompts that guide the model’s end-to-end segmentation by offering a comprehensive understanding of the image structure.
	Meanwhile, the high-frequency components $\mathbf{F}^H$ capture fine-grained details, particularly object boundaries.
	The high-frequency features $\mathbf{F}^H$ are fused with the features $\mathbf{F}^O$ from the CBOM encoder and $\mathbf{F}^K$	from the FDTIM. 
	This fusion combines complementary information from all three feature maps, enhancing the model's ability to accurately segment camouflaged objects, especially those with subtle or complex patterns. Finally,  the resulting fused feature map $\mathbf{F}^T$ is passed to the frozen decoder for UCIS task.

	\subsection{Channel Balance Optimization Module}
	
\begin{figure}[!t]
		\centering
		\includegraphics[width=0.49\textwidth]{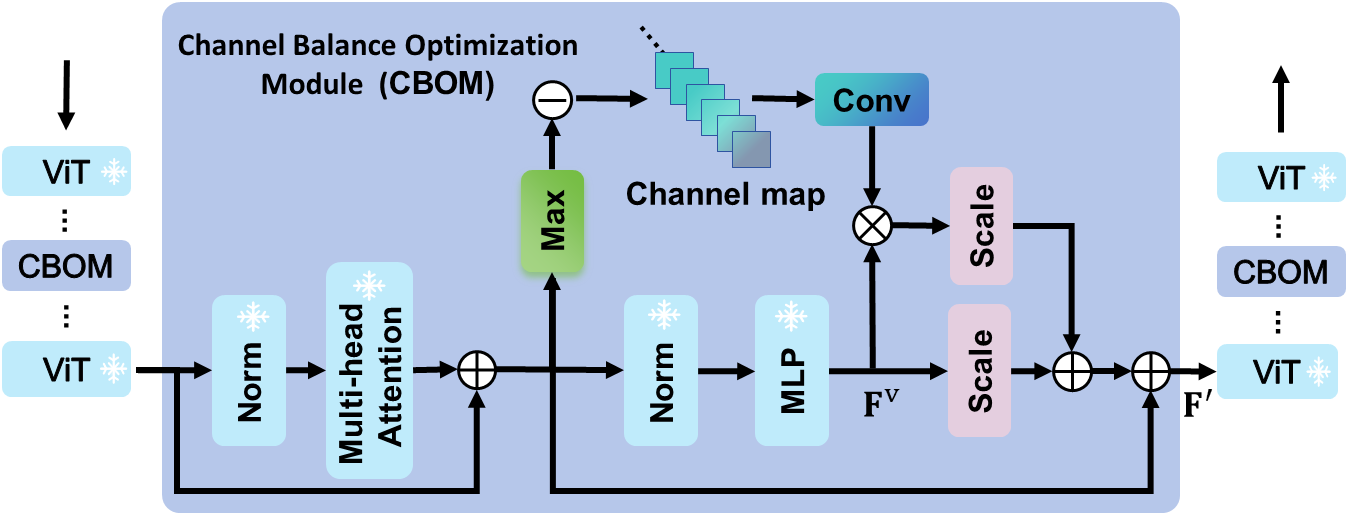}
		\caption{The Channel Balance Optimization Module. In the CBOM, the original VIT part remains frozen, while the channel properties are adjusted to mitigate the chromatic discrepancies and biases inherent in underwater images.}
		\label{CBOM}
	\end{figure}

	Typically, under ideal conditions devoid of any color bias, the average luminance of the red, green, and blue channels in an image should be approximately equal \cite{buchsbaum1980spatial}. 
	This assumption has been effectively utilized to enhance the visibility of images obscured by fog \cite{ju2021ide} and to mitigate challenges such as white balance distortion and low visibility in underwater images \cite{fan2024see,AN2024107219}. 
	Consequently, incorporating CBOM designed to eliminate color discrepancies and biases between channels in underwater images into SAM encoder is anticipated to enhance the model's feature extraction efficiency and segmentation accuracy in underwater environments. A detailed illustration of CBOM is illustrated in Fig. \ref{CBOM}.

	The absorption and scattering of light in underwater environments lead to varying degrees of attenuation across the red, green, and blue channels, thereby introducing channel imbalances in underwater images. 
	These imbalances are further propagated to the feature maps, where regions with the least attenuation are represented by pixels with the highest intensity values in their respective channels. 
	To prevent imbalances in the feature maps, we extract the maximum values $M_{ij}$ from each channel at every spatial location in the feature map $\mathbf{F}\in \mathbb{R}^{H\times W\times C}$ and use them as reliable reference points:
	
	\begin{equation}
		\label{C1}
		M_{ij}=max\left ( F_{ij0},F_{ij1}, \dots, F_{ijk}, \dots \right ),
	\end{equation}
	where $i$, $j$, $k$ are the indices of height, width, and channel, respectively. $F_{ijk}$ represents the intensity value at position $(i,j)$ in channel $k$. Thus, the channel reference matrix $\mathbf{M} \in \mathbb{R}^{H\times W\times 1}$ is constructed.
	
	To compare and quantify color bias, the average value of each channel is calculated and considered as the representative standard for that channel. 
	For each channel $k$, the average value $ \mu _{k}$ is calculated as:
	\begin{equation}
		\label{c3}
		\mu _{k}=\frac{1}{H\times W} \sum_{i=0}^{H-1} \sum_{j=0}^{W-1} F_{ijk}, 
	\end{equation}
	where $0\leq k\leq C-1$.
	Similarly, the standard value $ \mu _{r}$ of the reference channel $\mathbf{M}$ is calculated as:
	\begin{equation}\label{c4}
		\mu _{r}=\frac{1}{H\times W} \sum_{i=0}^{H-1} \sum_{j=0}^{W-1} M_{ij}.
	\end{equation}
	
	\noindent By comparing the discrepancies between these two sets of standard values, an estimation of the color bias $D_k$ for each channel $k$ can be defined as:
	\begin{equation}\label{c5}
		D_{k}= \mu _{r}-\mu _{k}.   
	\end{equation}
	
	\noindent Thus, $\mathbf{D} = \left[ D_{0}, D_{1}, \dots, D_{C-1} \right] $ distinctly describes the degree of deviation of each feature channel relative to the reference channel. The final channel bias map $\mathbf{D}^{\prime}$ is then obtained as:
	\begin{equation}
		\label{c6}
		\mathbf{D}^{\prime} =\sigma(Conv_{1}(GELU(Conv(\mathbf{D})))),   
	\end{equation}
	where $\sigma(\cdot)$ denotes the Sigmoid activation function, $Conv_{1}$ represents a $1\times 1$ convolution, and $GELU$ is the GELU activation function. The channel bias maps $\mathbf{D}^{\prime}$ are then element-wise multiplied with the feature maps $\mathbf{F}^{V}$ extracted by the original ViT. This operation is further balanced by a weighting factor $\lambda$, which controls the contribution of the corrected and original features. The resulting feature maps $\mathbf{F}^{\prime}$ provide a more accurate and robust feature representation for subsequent processing stages. Formally, it is expressed as:
	\begin{equation}
		\label{c7}
		\mathbf{F}^{\prime}=\lambda\mathbf{F}^{V}\odot \mathbf{D}^{\prime}+\left ( 1-\lambda \right ) \mathbf{F}^{V},   
	\end{equation}
	where $\odot$ represents element-wise multiplication operation.

	\subsection{Multi-scale Feature Frequency Aggregation Module}
	The SAM requires the user to provide foreground points, bounding boxes, or masks to guide the model's segmentation. Accordingly, it is essential to generate some prompts to feed into the SAM's decoder to obtain the camouflage instance segmentation masks. 
	Several methods have been proposed for generating such prompts\cite{chen2024rsprompter,lian2024diving,zhang2024uv}, including creating masks for all objects in an image for subsequent classification, using object bounding boxes from detectors as prior prompts, and so on.
	We design MFFAM as shown in Fig. \ref{framework} to directly predict the prompt embedding of camouflaged objects.
	In the frequency domain, low-frequency components primarily contain the color and content information of an image, while high-frequency components are mainly responsible for texture and detail information\cite{cong2024srnsd}. In the context of camouflaged images, an overabundance of texture and detail information can result in the model misidentifying objects.
	Therefore, during prompt generation, we only utilize the low-frequency components to ensure that the extracted features more accurately reflect the global contextual information. Simultaneously, high-frequency information representing finer details is further fused into the feature map to enhance the boundary information of the camouflaged objects.
		
	The output features $\mathbf {F}^{\prime}$ from the CBOM undergo DWT, which decomposes them into low-frequency and high-frequency components as follows:
	\begin{equation}
		\label{m1}
		LL_s,LH_s,HL_s,HH_s=DWT\left ( \mathbf {F}^{\prime}_s \right ),   
	\end{equation}
	where $s=\{1, 2, 3, 4\}$ represents the four distinct feature vectors derived from the CBOM output, $LL_s$ denotes the low-frequency component, and $LH_s,HL_s,HH_s$ correspond to high-frequency components in the vertical, horizontal, and diagonal directions, respectively.
	
	The low-frequency components are concatenated to enhance the representational capacity of the feature, and their dimensions are aligned to facilitate subsequent processing steps in the pipeline. It can be formulated as:
	\begin{equation}\label{m2}
		\mathbf{F}_{}^{LL} =Up\left ( Conv_1\left ( cat\left ( {LL}_1,  {LL}_2, {LL}_3, {LL}_4 \right )  \right )  \right ), 
	\end{equation}
	where $cat\left (\cdot \right) $ is the concatenation operation, $Conv_1$ is the $1\times1$ convolution used to adjust the number of channels, $Up$ is the upsampling operation that restores the  original space dimensions of the input feature.
	We then employ $3\times3$ convolutions to extract features and incorporate residual connections to enhance the network's learning capabilities and stability, which can be denoted as follows:
	\begin{equation}
		\label{m3}
		\mathbf{F}_{}^{L}= 
		Conv\left ( \mathbf{F}_{}^{LL} \right )  +\mathbf{F}_{}^{LL}.  
	\end{equation}
	Considering the scale variability of the camouflaged objects, we apply multi-scale transposed convolutional layers for $2\times$ and $4\times$ upsampling of the feature $\mathbf{F}_{}^{L}$. Additionally, a max pooling operation is applied for 1/2 and 1/4 downsampling of the feature $\mathbf{F}_{}^{L}$. These features, along with the original feature $\mathbf{F}_{}^{L}$, are then fed into the Region Proposal Network (RPN) header \cite{ren2016faster}, comprising five distinct scale representations.

	The high-frequency components encompass abundant details, particularly in terms of edge and contour features \cite{li2023feature}.
	Here, we utilize Eq. (\ref{m4}) to extract the magnitude information across various directions to capture the fine structures. Meanwhile, the energy distribution of the high-frequency coefficient is evaluated in Eq. (\ref{m5}) to gain insights into the internal dynamics of the features.
	
	\begin{equation}\label{m4}
		\begin{matrix}
			H_s^{abs}=
			\left | LH_{s}\right | +\left | HH_{s}\right | +\left | HL_{s}\right |,
		\end{matrix}
	\end{equation}
	
	\begin{equation}\label{m5}
		H_s^{sqrt}=\sqrt{LH_{s}^2+HH_{s}^2+HL_{s}^2} . 
	\end{equation}
	A comprehensive high-frequency information $H_s$ can be obtained as follows:
	\begin{equation}\label{m6}
		H_s=H_s^{abs}+H_s^{sqrt}.
	\end{equation}
	
	\noindent To ensure the comparability of high-frequency information across different feature levels, normalization is first applied following the reconstruction of the high-frequency information, which can be expressed by:
	\begin{equation}\label{m7}
		H_{all}=\sum \left ( H_s\times \frac{H_s}{\sum H_s}  \right ). 
	\end{equation}
	The final high-frequency feature map $\mathbf{F}_{}^{H}$ is obtained as below:
	\begin{equation}\label{m8}
		\mathbf{F}_{}^{H}= Conv\left ( Conv_1\left ( Up\left (  H_{all}\right )  \right )  \right ),
	\end{equation}
	where $Up$ is the upsampling operation.
	
	These specific high-frequency details are also superimposed on the original image features $\mathbf{F}^O$ obtained from the CBOM encoder to obtain more detailed and comprehensive features:
	\begin{equation}\label{m9}
		\mathbf{F}_{}^{O1}= \mathbf{F}_{}^{O}+\mathbf{F}_{}^{H}.
	\end{equation}
	
	\subsection{Frequency Domain Truth Integration Module}
	In camouflaged scenes, objects often leverages the color, texture, shape, and other characteristics of the surrounding environment to camouflage itself, which poses significant challenges for segmentation.
	In spatial domain, instance features are blended with those of the background. Therefore, adopting frequency domain for analysis may bring more possibilities for segmenting camouflaged objects. 
	Frequency domain processing techniques have achieved significant breakthroughs in tasks such as identifying fake images \cite{frank2020leveraging}, enhancing low-light remote sensing images \cite{yao2024spatial}.
	We propose FDTIM, designed to identify and filter out deceptive information in the frequency domain, protecting the real information from confusion while enhancing the learning of camouflaged object features.

	Applying a discrete two-dimensional Fourier transform to the original input image $x \in \mathbb{R} ^{M\times N\times 3} $ converts it from the spatial domain to the frequency domain, yielding the frequency spectrum $f\left ( u,v\right )$:
	\begin{equation}\label{e1}
		f\left ( u,v\right ) =
		\sum_{m=0}^{M-1} \sum_{n=0}^{N-1} 
		x\left ( m,n \right ) 
		\cdot e^{-j\cdot 2\pi \left ( \frac{um}{M} +\frac{vn}{N}  \right ) } ,
	\end{equation}
	where $j$ is the imaginary unit, $u$ and $v$ are the row and column coordinates in the frequency domain, respectively.
	Equivalently, the frequency spectrum  $f\left ( u,v\right )$ can also be represented as:
	\begin{equation}\label{e2}
		f\left ( u,v\right )=a\left ( u,v\right )+j\cdot b\left ( u,v\right ),
	\end{equation}
	where $a\left ( u,v\right )$ represents the real part, $b\left ( u,v\right )$ represents the imaginary part.
	The amplitude information $A\left ( u,v \right )$ at different frequencies is 
	\begin{equation}\label{e3}
		A\left ( u,v \right ) =
		\left |  f\left ( u,v\right )\right | =
		\sqrt{{a^2\left ( u,v\right )+ b^2\left ( u,v\right )} } .
	\end{equation}
	
	\noindent It reflects the intensity or prominence of that frequency component in the image. When a particular frequency component is dominant or frequently present in the image, its corresponding amplitude $A\left ( u,v \right )$ will be significantly increased.	
	
	Camouflaged objects in the image often resemble their surrounding environment, manifesting as frequency components with larger amplitudes. This distinctive amplitude offers a novel approach to identifying and removing camouflaged features, enabling the extraction of camouflaged information while preserving the underlying real content.
	Specifically, it is achieved by filtering the spectrum to isolate the top $K$ highest frequency components as shown below: 
	\begin{equation}\label{e4}
		f^{\prime} \left ( u,v\right )=
		\left \{ f \left ( u,v\right )|u,v\notin A_K \left ( u,v\right )\right \} ,
	\end{equation}
	where $ A_K \left ( u,v\right )$ is the top $K$ largest amplitude values, and $f^{\prime} \left ( u,v\right )$ is the filtered spectrum. Subsequently, these separated frequency components are reconstructed back into the spatial domain using the inverse Fourier transform, removing disruptive features and restoring the image's authenticity. The reconstructed image $x^{\prime} \left ( m,n \right )$ is expressed as:	
	\begin{equation}\label{e5}
		x^{\prime} \left ( m,n \right )  =
		{\frac{1}{MN} \sum_{u=0}^{M-1} \sum_{v=0}^{N-1} 
			f^{\prime} \left ( u,v\right )
			\cdot e^{j\cdot 2\pi \left ( \frac{um}{M} +\frac{vn}{N}  \right ) } },
	\end{equation}
	which is fed into frozen SAM encoder to obtain more authentic features $\mathbf{F}^K$.
	
	Based on features $\mathbf{F}^{O1}$ which have been previously superimposed with high-frequency information, we perform a subtraction operation between these two feature maps, followed by an element-wise multiplication with the feature map to extract and enhance the genuine information while suppressing the influence of camouflaged features. The final truth features $\mathbf{F}^T$ are formulated as:
	\begin{equation}\label{e6}
		\mathbf{F}^T=
		\mathbf{F}^{O1}\odot \sigma\left (  Conv\left ( \mathbf{F}^{O1} - \mathbf{F}^K \right )\right ).    
	\end{equation}
	Truth features $\mathbf{F}^T$, resulting from the integration of the outputs from the CBOM, FDTIM, and MFFAM modules, are then input into the frozen mask decoder to generate the final segmentation results.
	
		\begin{table}[!t]
		\vspace{-0.25cm}
		\renewcommand\arraystretch{1.5}
		\caption{Quantitative comparisons with state-of-the-arts methods on our UCIS4K datasets, where \textbf{bold} denotes the best performance, and \underline{underlined} denotes the second best. 
			\label{quantitative_comparisons_UCIS}} 
		\centering
		\begin{tabular}{c|c|c|ccc}
			\Xhline{1pt}
			\multirow{2}{*}{Methods}& \multirow{2}{*}{Pub’Year}&\multirow{2}{*}{Backbone}  & \multicolumn{3}{c}{UCIS4K} \\
			\cline{4-6}
			
			&& & $\text{AP}$& $\text{AP}_{50}$& $\text{AP}_{75}$ \\
			\hline
			OSFormer\cite{pei2022osformer}&ECCV’22&ResNet-50&47.7 & 71.2 & 52.1\\
			OSFormer\cite{pei2022osformer}&ECCV’22&ResNet-101&49.2 & 71.3 & 54.4       \\  
			CE-OST\cite{nguyen2023ost}&MAPR’23&ResNet-50&48.5 & 71.8 & 54.1\\
			CE-OST\cite{nguyen2023ost}&MAPR’23&ResNet-101&50.0 &\underline{73.1}& 54.8\\
			DCNet\cite{luo2023camouflaged}&CVPR’23&ResNet-50&50.5 & 69.5 & 54.9 \\
			DCNet\cite{luo2023camouflaged}&CVPR’23&ResNet-101&\underline{50.7} & 69.7 & \underline{55.9} \\
			Watermask\cite{lian2023watermask}&ICCV’23&ResNet-50&41.5 &66.6 &45.0\\
			Watermask\cite{lian2023watermask}&ICCV’23&ResNet-101&44.4 &69.2& 48.6\\
			Mask2Former\cite{cheng2022masked}&CVPR’22&ResNet-50&49.0 & 69.6 & 53.5\\ 
			Mask2Former\cite{cheng2022masked}&CVPR’22&ResNet-101&49.7 & 70.0 & 54.6\\

			SAM+mask\cite{kirillov2023segment}&ICCV’23&VIT-H&34.5 &60.8 &35.6\\
			SAM+bbox\cite{kirillov2023segment}&ICCV’23&VIT-H&40.4 &63.9 &43.3\\

			SAM2\cite{ravi2024sam}&-’24&Hiera-Large&11.6&13.9&12.6\\
			\hline
		
			UCIS-SAM&-&VIT-H&\textbf{54.0}&\textbf{77.8}&\textbf{59.6}\\

			\Xhline{1pt}
		\end{tabular}
		\vspace{-0.05cm}
	\end{table}

	\section{Experiments}\label{section55}
	\subsection{Datasets}
	To validate the effectiveness of our UCIS-SAM model, we conducted extensive experiments using four datasets, categorized into three groups:
	\subsubsection{Underwater Camouflaged Instance Segmentation Datasets} Our UCIS4K dataset contains 3,953 underwater camouflaged images with instance-level annotations, divided into 2,967 training images and 986 testing images.	
	\subsubsection{Underwater Instance Segmentation Datasets} The UIIS\cite{lian2023watermask} is an underwater image instance segmentation dataset, containing 3,937 training images and 691 testing images. 
	\subsubsection{Camouflaged Instance Segmentation Datasets} The COD10K\cite{fan2020camouflaged} dataset contains 3040 camouflaged images with instance-level annotations for training and 2026 images for testing. As a supplementary dataset, NC4K\cite{lv2021simultaneously} includes 4121 testing camouflaged images to evaluate the model's generalization capability. While both datasets feature a small number of underwater camouflaged images, they primarily focus on terrestrial camouflaged organisms.

	\subsection{Evaluation Metrics \& Experimental Settings}
	In this research, we focus on segmenting instances within camouflaged images. The standard mask AP metrics \cite{lin2014microsoft}, including AP, AP50, and AP75, are employed to evaluate the performance of our model. These metrics are consistent with the evaluation criteria commonly used in the field of class-agnostic camouflaged instance segmentation.

	The UCIS-SAM model is trained on 2 NVIDIA GeForce RTX 4090 GPUs with a batch size of 2, employing the AdamW optimizer with a base learning rate of $1\text{e-}4$ for 30 epochs.
	We implement a Cosine Annealing scheduler \cite{loshchilov2016sgdr} with a linear warm-up strategy to gradually increase the learning rate before decaying it.
	During the training phase, the backbone network employs the Vision Transformer (ViT-H), where all layers are frozen except for the previously mentioned modules. The hyperparameter $\lambda$ is set to 0.2 in the CBOM, and the hyperparameter $K$ is set to 1000 in the FDTIM empirically.	
	
		\begin{table}[!b]
		\vspace{-0.15cm} 
		\renewcommand\arraystretch{1.5}
		\caption{Quantitative comparisons with state-of-the-art methods on UIIS datasets, where \textbf{bold} denotes the best performance, and \underline{underlined} denotes the second best. }
		\label{quantitative_comparisons_UIIS}
		\centering
		\begin{tabular}{c|c|c|ccc}
			\Xhline{1pt}
			\multirow{2}{*}{Methods}& \multirow{2}{*}{Pub’Year}&\multirow{2}{*}{Backbone}  & \multicolumn{3}{c}{UIIS} \\
			\cline{4-6}
			
			&& & $\text{AP}$& $\text{AP}_{50}$& $\text{AP}_{75}$ \\
			\hline
			Watermask\cite{lian2023watermask}&ICCV’23&ResNet-50&23.3 &39.7 &24.8      \\
			Watermask\cite{lian2023watermask}&ICCV’23&ResNet-101& 25.6& 41.7 &27.9     \\
			Mask2Former\cite{cheng2022masked}&CVPR’22&ResNet-50&36.3 & 56.3 & 38.8      \\ 
			Mask2Former\cite{cheng2022masked}&CVPR’22&ResNet-101& 36.3& 56.5 & 38.3       \\
		
			OSFormer\cite{pei2022osformer}&ECCV’22&ResNet-50&36.1 & 57.9 & 38.1   \\
			OSFormer\cite{pei2022osformer}&ECCV’22&ResNet-101&\underline{36.7} & \underline{58.1} & 38.5        \\  
			CE-OST\cite{nguyen2023ost}&MAPR’23&ResNet-50&35.9&57.3&37.6 \\
			CE-OST\cite{nguyen2023ost}&MAPR’23&ResNet-101&36.4& 57.6& 38.7\\
			DCNet\cite{luo2023camouflaged}&CVPR’23&ResNet-50&21.7 & 33.8 & 22.6   \\
			DCNet\cite{luo2023camouflaged}&CVPR’23&ResNet-101&23.3 &36.0 & 24.2\\

			SAM+mask\cite{kirillov2023segment}&ICCV’23&VIT-H&25.1 &50.9 &21.7  \\
			SAM+bbox\cite{kirillov2023segment}&ICCV’23&VIT-H&36.2 &57.1 &\underline{39.5}  \\

			SAM2\cite{ravi2024sam}&-’24&Hiera-Large&17.9 &23.6 &19.7 \\
			
			\hline
			
			UCIS-SAM&-&VIT-H&\textbf{39.0} &\textbf{61.0} &\textbf{41.6}\\
			
			\Xhline{1pt}
		\end{tabular}
	\end{table}

	\begin{figure*}[!t]
		\centering
		\vspace{-0.25cm} 
		\includegraphics[width=0.98\textwidth]{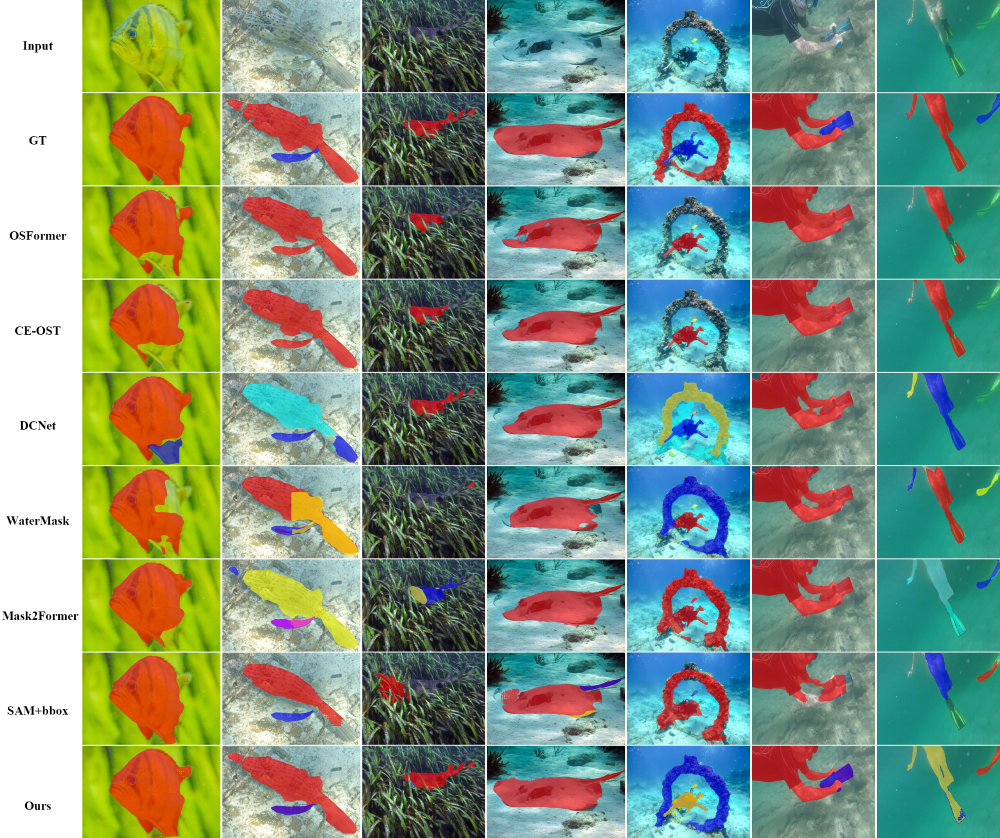}
		\vspace{-0.15cm} 
		\caption{Comparison of results with other instance segmentation methods on UCIS4K and UIIS dataset.
			From top to bottom: the original image is followed by ground truth and results of OSFormer \cite{pei2022osformer}, CE-OST\cite{nguyen2023ost}, DCNet \cite{luo2023camouflaged}, WaterMask \cite{lian2023watermask}, Mask2Former \cite{cheng2022masked}, SAM+bbox \cite{kirillov2023segment} and our UCIS-SAM. Each camouflaged instance is represented by a unique color. The first 4 columns are from our UCIS4K dataset, and the last 3 columns are from the UIIS dataset. }
		\label{result10}
		\vspace{-0.05cm}
	\end{figure*}

	\begin{figure*}[!t]
		\vspace{-0.15cm}
		\centering
		\includegraphics[width=0.98\textwidth]{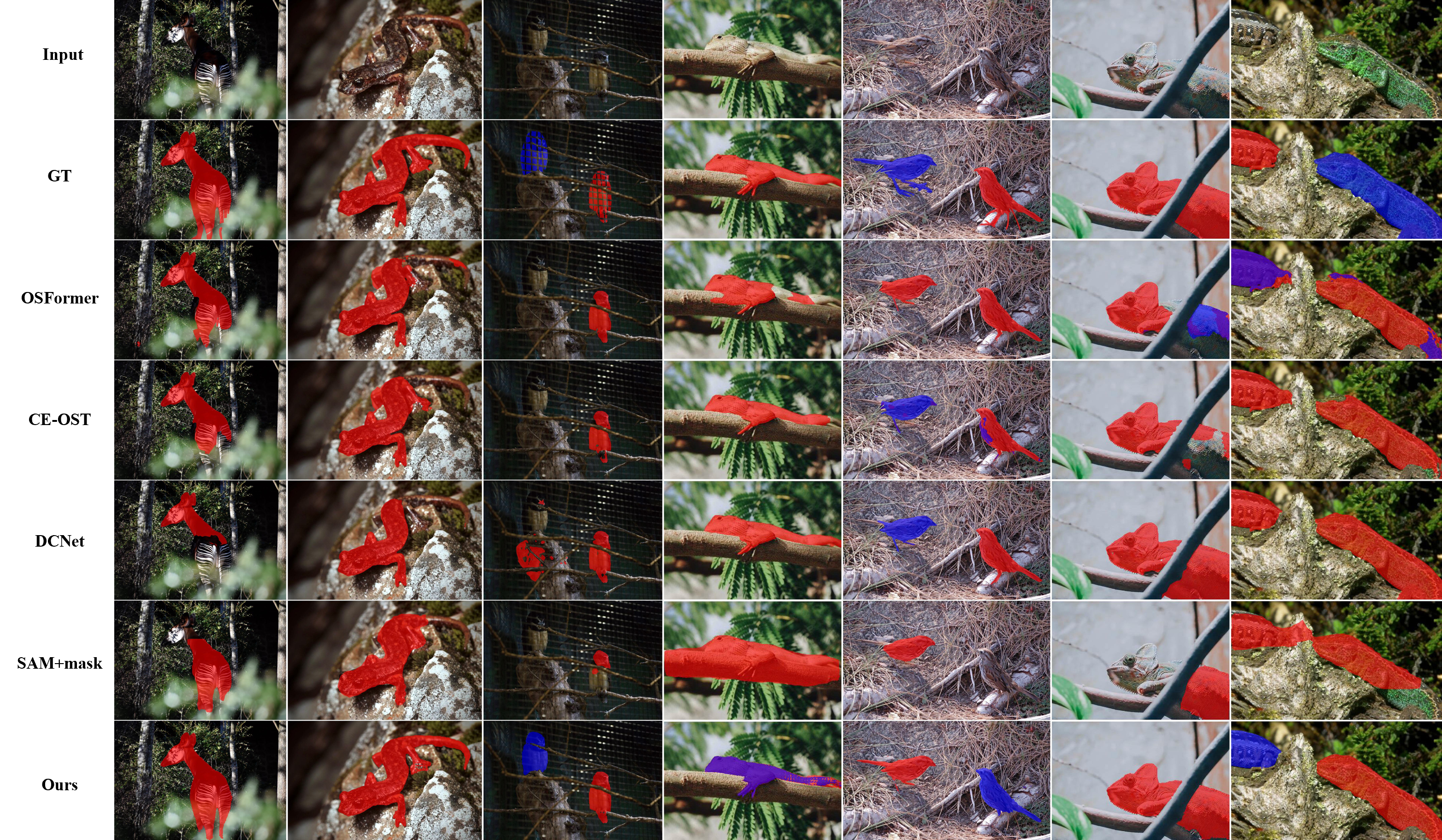}
		\vspace{-0.15cm} 
		\caption{Comparison with other CIS methods on COD10K and NC4K dataset. From top to bottom: the original image is followed by ground truth and results of OSFormer \cite{pei2022osformer}, CE-OST\cite{nguyen2023ost}, DCNet \cite{luo2023camouflaged},  SAM+mask \cite{kirillov2023segment} and our UCIS-SAM. Each camouflaged instance is represented by a unique color. The first 3 columns are from COD10K dataset, and the last 4 columns are from NC4K dataset. UCIS-SAM also demonstrates comparable performance.}
		\label{result22}
	\end{figure*}
	
	\begin{table*}[!t]
		\vspace{-0.15cm}
		\renewcommand\arraystretch{1.5}
		\caption{Quantitative comparisons with state-of-the-art methods on COD10K and NC4K datasets, where \textbf{bold} denotes the best performance, and \underline{underlined} denotes the second best.
			\label{quantitative_comparisons_COD10K}} 
		\centering
		
		\begin{tabular}{c|c|c| p{0.8cm}<{\centering} p{0.6cm}<{\centering} p{0.8cm}<{\centering}|p{0.8cm}<{\centering} p{0.6cm}<{\centering} p{0.8cm}<{\centering}}
			\Xhline{1pt}
			\multirow{2}{*}{Methods}& \multirow{2}{*}{Pub’Year}&\multirow{2}{*}{Backbone}  & \multicolumn{3}{c|}{COD10K}& \multicolumn{3}{c}{NC4K} \\
			\cline{4-9}
			&&&$\text{AP}$& $\text{AP}_{50}$& $\text{AP}_{75}$&$\text{AP}$& $\text{AP}_{50}$& $\text{AP}_{75}$\\
			\hline
		
			OSFormer\cite{pei2022osformer}&ECCV’22&ResNet-50&41.0 &71.1& 40.8 &42.5& 72.5& 42.3\\
			OSFormer\cite{pei2022osformer}&ECCV’22&ResNet-101&42.0 &71.3 &42.8 &44.4& 73.7& 45.1        \\  
			CE-OST\cite{nguyen2023ost}&MAPR’23&ResNet-50&41.6&70.7&42.3&42.4&71.4&42.6\\
			CE-OST\cite{nguyen2023ost}&MAPR’23&ResNet-101&43.2&72.2&44.1&45.1&74.0&46.4\\
			UQFormer\cite{dong2023unified}&ACM MM’23&ResNet-50&45.2&71.6&46.6&47.2&74.2&49.2\\
			UQFormer\cite{dong2023unified}&ACM MM’23&ResNet-101&45.5&71.8&47.9&50.1&76.8&52.8\\ 
			DCNet\cite{luo2023camouflaged}&CVPR’23&ResNet-50&45.3 &70.7 &47.5&52.8 &77.1& 56.5  \\
			DCNet\cite{luo2023camouflaged}&CVPR’23&ResNet-101&46.8 &72.9 &49.0&\underline{54.0}&78.3&\underline{58.0}\\
		
			GLNet\cite{li2024camouflaged}&IEEE Signal Process Lett’24&P2T\cite{wu2022p2t}&\underline{49.3}&\underline{77.9}&\underline{52.7}&53.4&\underline{81.0}&57.9\\

			SAM+mask\cite{kirillov2023segment}&ICCV’23&VIT-H&21.8 &47.9 &17.1&27.6 &58.1&22.6  \\
			SAM+bbox\cite{kirillov2023segment}&ICCV’23&VIT-H&30.9 &54.7 &31.5&33.8&59.5 &33.7  \\
		
			SAM2\cite{ravi2024sam}&-’24&Hiera-Large&10.6& 13.2& 11.8&8.8&10.3&9.6\\
			\hline
		
			UCIS-SAM&-&VIT-H&\textbf{50.7} &\textbf{78.7} &\textbf{55.1}&\textbf{56.8} &\textbf{83.3} &\textbf{62.7}\\
			
			\Xhline{1pt}
		\end{tabular}
		\vspace{-0.05cm}
	\end{table*}

	\subsection{Experimental Results}
	We first conducted experiments on the proposed UCIS4K dataset. Since it is the first dataset for underwater camouflaged instance segmentation, we then compared our model with the state-of-the-art methods on UIIS, COD10K and NC4K datasets to further verify the generalization ability of our model.

    We compare the performance of UCIS-SAM with state-of-the-art methods, including CIS approaches such as OSFormer \cite{pei2022osformer}, CE-OST \cite{nguyen2023ost}, and DCNet \cite{luo2023camouflaged}, underwater instance segmentation (UIS) methods like WaterMask \cite{lian2023watermask}, and general instance segmentation (GIS) techniques, including Mask2Former \cite{cheng2022masked} and the SAM series \cite{kirillov2023segment,ravi2024sam}. For SAM2, $32^{2}$ points are uniformly generated across the image to function as input prompts \cite{lian2024evaluation}, corresponding to the `automatic' setting.
    
	\subsubsection{UCIS4K dataset}	
	All the compared methods are trained and evaluated on our UCIS4K dataset using their officially released code. 	
	The quantitative results are presented in Table \ref{quantitative_comparisons_UCIS}.
	Our proposed UCIS-SAM model outperforms the compared state-of-the-art methods in the field of underwater camouflaged instance segmentation. Specifically, UCIS-SAM achieves improvements of 3.3, 4.7, and 3.7 in terms of $\text{AP}$, $\text{AP}_{50}$, and $\text{AP}_{75}$, respectively, compared to the second-best performing method. These results underscore the superior capability of UCIS-SAM in accurately segmenting camouflaged objects in challenging underwater environments, effectively addressing the unique challenges posed by these settings.
	Compared to UIS methods such as Watermask, UCIS-SAM demonstrates substantial improvements, achieving increases of 9.6, 8.6, and 11.0 in $\text{AP}$, $\text{AP}{50}$, and $\text{AP}{75}$, respectively. These results emphasize the distinct advantages of UCIS-SAM in camouflage segmentation.
	For SAM-based models, the lack of domain-specific knowledge in the underwater and camouflage contexts within the encoder of SAM results in a significant performance gap for variants like SAM+mask and SAM+bbox. This highlights the fact that, while large pre-trained models exhibit impressive generalization capabilities, the integration of domain-specific expertise is essential for optimizing performance in specialized tasks.
	
	A comparative visualization of our method against other tested approaches is shown in the first 4 columns of Fig. \ref{result10}, where the backbones of the latter are selected based on their highest performance metrics on the UCIS4K dataset. Our UCIS-SAM method consistently outperforms all other approaches, yielding results that most closely align with the ground truth.
	In scenarios where the instance's color and texture closely resemble the background (column 1) or in cases of partially occluded camouflaged instances (the fish’s head in column 2, the fish’s tail and head in column 3), UCIS-SAM exhibits its semantic-level understanding by fully segmenting the instances, in contrast to other methods that either fail or provide partial segmentation due to background interference. 
	It is largely attributed to the FDTIM, which effectively mitigates the impact of camouflaged features and enhances the differentiation between instances and their backgrounds, enabling the model to better comprehend instance semantics and improve segmentation accuracy.
	In cases involving camouflaged instances with ambiguous boundaries and underwater lighting interference (column 4), UCIS-SAM excels in capturing subtle boundary differences and achieving precise segmentation, unaffected by light speckles. This is made possible by CBOM and MFFAM, which effectively handle the challenges of underwater environments and enhance the boundary and fine-grained details of objects, ensuring reliable segmentation under complex environmental conditions.

\subsubsection{UIIS Dataset}
We further evaluate the performance of UCIS-SAM on the UIIS dataset, with all methods trained and evaluated on this dataset. As shown in Table \ref{quantitative_comparisons_UIIS}, our method shows notable improvements over state-of-the-art approaches.
Compared to the second-best method, UCIS-SAM improves by 2.3, 2.9, and 2.1 in $\text{AP}$, $\text{AP}_{50}$, and $\text{AP}_{75}$, respectively.
The visual results in columns 5 to 7 of Fig. \ref{result10} clearly show that our segmentation results closely align with the ground truth and effectively adapt to underwater color distortion (columns 5, 7).
These results highlight the model's ability to handle the unique challenges of underwater environments, with the proposed CBOM playing a key role in addressing issues such as color distortion and color imbalance. UCIS-SAM shows strong learning capabilities in the underwater domain, achieving more accurate instance segmentation despite the complexities of underwater images.

\subsubsection{COD10K and NC4K Dataset}	
Several state-of-the-art CIS methods and SAM-based models are selected for comparison with our UCIS-SAM. All models are trained on COD10K training set and evaluated on  COD10K and NC4K testing sets.
Quantitative results in Table \ref{quantitative_comparisons_COD10K} show that UCIS-SAM outperforms other methods on both COD10K and NC4K datasets, with improvements of 1.4, 0.8, and 2.4 in $\text{AP}$, $\text{AP}_{50}$, and $\text{AP}_{75}$ on COD10K, and 2.8, 2.3, and 4.7 on NC4K.
\begin{table}[!t]
	\vspace{-0.25cm}
	\renewcommand\arraystretch{1.5}
	\caption{Ablation studies on the impact of different components in UCIS-SAM model. ``All'' refers to CBOM, MFFAM, and FDTIM.}
	\label{ablation}
	\centering
	\begin{tabular}{c|ccc}
		\Xhline{1pt}
		Architectures Design& $\text{AP}$&$\text{AP}_{50}$ & $\text{AP}_{75}$ \\
		\hline
		
		w/o CBOM  & 52.1(-1.9)&75.8(-2.0)&56.7(-2.9)\\
		w/o MFFAM  & 51.2(-2.8) &75.8(-2.0) &56.2(-3.4)\\
		w/o FDTIM  & 52.3(-1.7)&76.1(-1.7)&57.0(-2.6)\\
		w/o ALL &50.4(-3.6) &74.9(-2.9) &54.0(-5.6)\\
		UCIS-SAM &54.0&77.8&59.6\\
		\Xhline{1pt}
	\end{tabular}

\end{table}

\begin{figure}[!t]
	\setlength{\abovecaptionskip}{-2pt}
	\centering
	\includegraphics[width=0.4\textwidth]{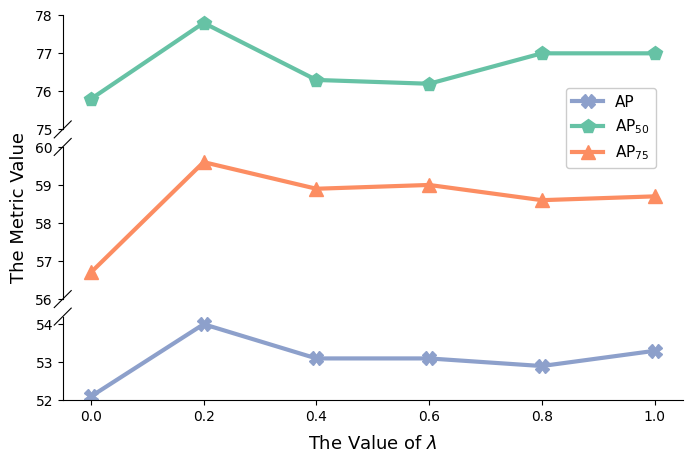}
	\vspace{-0.15cm} 
	\caption{The fusion strategy with varying values of $\lambda$ in CBOM.}
	\label{lambda}
\end{figure}

We further conducted a visual evaluation of UCIS-SAM's performance, comparing it with other open-source methods, as shown in Fig. \ref{result22}. The results clearly demonstrate the unparalleled performance of UCIS-SAM.
It effectively integrates contextual information, enhancing its ability to accurately understand and segment instances, thereby ensuring precise delineation without compromising the integrity of the object (columns 2, 4). It exhibits robust performance even in the presence of partial occlusion or truncation (columns 1, 6), and excels in capturing fine details even when objects are significantly occluded (column 3). In multi-object scenarios, UCIS-SAM demonstrates exceptional discriminative ability, effectively preventing overlap and ambiguity between objects, ensuring independent and precise segmentation of each instance (columns 5, 7).
These results underscore the model's ability to accurately delineate object boundaries, maintain robustness in challenging conditions like occlusion or truncation, and ensure precise segmentation in multi-object contexts, highlighting its potential for broader application in other scenarios.
The exceptional segmentation performance of UCIS-SAM can be primarily attributed to the application of SAM. In tackling domain-specific camouflage challenges, FDTIM effectively distinguishes easily confusable camouflage features, while MFFAM enhances ambiguous boundaries. These components together enable UCIS-SAM to efficiently address the significant challenges posed by camouflage.
Further visualizations are available in the supplementary materials.

	\subsection{Ablation Studies}\label{section5}
	To investigate the effect of our core designs, we perform a series of studies on the UCIS4K dataset. 
	
	\subsubsection{Analysis of CBOM}
	The CBOM block is removed to validate its performance, employing the unmodified SAM encoder directly in the model.
	As demonstrated in Table \ref{ablation}, the model's performance on the $\text{AP}$, $\text{AP}_{50}$, and $\text{AP}_{75}$ metrics is decreased by 1.9, 2.0, and 2.9, respectively. 
	It indicates that the CBOM is crucial for mitigating chromatic aberrations and color deviations in underwater environments.
	It enhances the model's ability to extract unique features from underwater images for more effective processing of these environments.

	As previously mentioned, the features in CBOM are fused using a parameter $\lambda$, which is governed by Eq. (\ref{c7}) to control the balance between the original feature map and the corrected feature map with the channel bias map. 
	We conduct experiments with different values of $\lambda$, selecting values at intervals of 0.2.
	As shown in Fig. \ref{lambda}, when $\lambda=0$, the channel bias map is not integrated, leading to a noticeable decrease in model performance.
	Conversely, when $\lambda =1$, the original feature map is not utilized, resulting in suboptimal model performance.
	A fusion strategy with a smaller weight of $\lambda =0.2$ improves the model's ability to process underwater images, which is beneficial for preserving more original image information while moderately incorporating adjustments from the channel bias map into the features.
	Consequently, it maintains image details and mitigates color bias and chromatic aberrations in underwater environments.

	\subsubsection{Analysis of MFFAM}
	In the MFFAM, the features processed through DWT are divided into two parts: the low-frequency components are fused and fed into the RPN head, while the high-frequency components are fused and then superimposed onto the feature maps generated by the CBOM encoder.
	We conduct the experiment where the features are directly fed into the RPN head without incorporating the low-frequency components extracted by DWT, and the fusion of high-frequency information is omitted.
	We conduct experiments where features are directly input into the RPN head without the low-frequency components in DWT, and the fusion of high-frequency information is omitted.
	According to the results from Table \ref{ablation}, the model's performances on the $\text{AP}$, $\text{AP}_{50}$, and $\text{AP}_{75}$ metrics are decreased by 2.8, 2.0, and 3.4, respectively. It suggests that feature fusion after DWT is crucial for improving the model's performance.	
	
	To evaluate the individual contribution of both the low-frequency and high-frequency components to the overall performance of the model, we carry out experiments by selectively removing either the high-frequency or low-frequency components.
	The results are presented in Table \ref{high low frequency}.
	In terms of the AP metric, the model's performance decreases by 2.1 when only the low-frequency components from DWT are used, and by 1.6 when only the high-frequency components are used.
	It indicates that high-frequency and low-frequency information play distinct roles. The high-frequency component primarily encompasses the local features and details of an image, while the low-frequency information encompasses the global structure of the image. 
	Confusing these two types of information can hinder the model's ability to accurately capture key features, which in turn affects its generalization capability and overall performance.
	Removing both high and low frequency components entirely from MFFAM would significantly degrade the model's performance, resulting in a 2.8 reduction.
	This demonstrates that the separation and independent processing of high and low frequency information are crucial for segmenting camouflaged objects.

	\begin{table}[!t]
		\vspace{-0.25cm}
		\renewcommand\arraystretch{1.5}
		\caption{High-frequency and low-frequency components in MFFAM.} 
		\label{high low frequency}
		\centering
		\begin{tabular}{cc|p{0.9cm}<{\centering} p{0.8cm}<{\centering} p{0.9cm}<{\centering}}
			\Xhline{1pt}
			Low-frequency&High-frequency&$\text{AP}$& $\text{AP}_{50}$& $\text{AP}_{75}$\\
			\hline		
			
			\XSolidBrush&\XSolidBrush& 51.2 &75.8 &56.2\\
			\XSolidBrush&\CheckmarkBold&52.4& 76.2&57.7\\
			\CheckmarkBold&\XSolidBrush&51.9 &75.7 &57.1\\
			
			\CheckmarkBold&\CheckmarkBold&54.0&77.8&59.6\\
			
			\Xhline{1pt}
		\end{tabular}
	\end{table}

	\begin{figure}[!t]
		\setlength{\abovecaptionskip}{-2pt}
		\centering
		\includegraphics[width=0.4\textwidth]{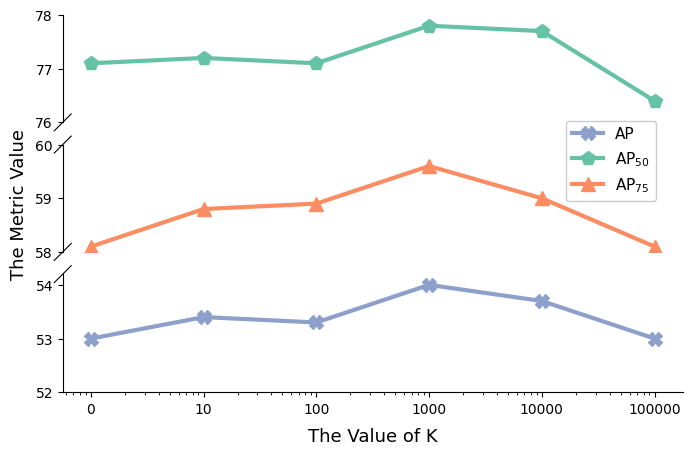}
		\vspace{-0.1cm}
		\caption{Strategies for selecting camouflaged components in FDTIM.}
		\label{K}
	\end{figure}

	\subsubsection{Analysis of FDTIM}
	According to the results presented in Table \ref{ablation}, the model's performance improves by 1.7, 1.7, and 2.6 in $\text{AP}$, $\text{AP}_{50}$, and $\text{AP}_{75}$ metrics, respectively, following the integration of FDTIM.
	This outcome demonstrates the efficacy of Fourier transform-based amplitude filtering in the elimination of camouflage information.
	It means that FDTIM has effectively reduced certain camouflaged features, thereby allowing the authentic features to be more prominently highlighted and maximized.

	The parameter $K$ as shown in Eq. (\ref{e4}) in FDTIM is used to filter the camouflaged components in an image. 
	Given that all input images are resized to $1024\times1024\times3$, there are over three million frequency components in the spectrum. 
	To find an appropriate value for $K$, experiments are carried out with different settings from 0 to 100,000 (approximately 1/30 of the frequency components).
	The experimental results in Fig. \ref{K} indicate that the model's performance gradually deteriorates as the parameter $K$ increases from 1,000 to 100,000.
	This trend suggests that larger values of $K$ may lead to the loss of some crucial information, impeding the model's ability to generalize effectively.
	Optimal performance is attained at $K=$1,000, which implies that an optimal balance is struck between the elimination of superfluous camouflaged components and the retention of essential features necessary for the model's discernment. 
	Conversely, when $K$ is reduced to 0, the model's performance deteriorates due to the insufficient removal of camouflaged features, thereby limiting its ability to distinguish between salient and spurious information. 
	Therefore, selecting an appropriate $K$ value is important for retaining the useful information required by the model.

	\subsubsection{Analysis of SAM}To further validate the contribution of the three proposed modules CBOM, MFFAM, and FDTIM to overall model performance, we conducted an ablation study removing all newly introduced modules and retaining only the SAM baseline model. The results are summarized in Table \ref{ablation}. Upon removal of these modules, the model’s $\text{AP}$, $\text{AP}_{50}$, and $\text{AP}_{75}$ decrease by 3.6, 2.9, and 5.6, respectively, with a significant performance drop. This suggests that the enhanced performance of UCIS-SAM is not solely attributable to the SAM baseline, but is significantly influenced by the synergistic effects of multiple modules. Moreover, the performance degradation observed after removing all three modules is more pronounced than the removal of any single module, further underscoring their critical role in boosting overall model performance. Therefore, it can be concluded that the improvement is not solely attributable to SAM, but rather to the combined effect of CBOM, MFFAM, and FDTIM.
	
	\subsection{Discussion \& Future Work}\label{section6}
	In this work, we proposed the first UCIS4K dataset for underwater camouflaged instance segmentation task. Since the underwater dataset is limited and images of camouflage characteristics are difficult to acquire, we will continuously expand and update the dataset with subsequent accumulation.    
    Moreover, to evaluate the proposed UCIS4K dataset, we devised the UCIS-SAM model for underwater scenes. Furthermore, it has also shown promising performance in other scenes by our experiments on some other dataset. We will optimize the architecture of the model and explore its possibilities in other challenging scenes in future work.

	\section{Conclusion}\label{section7}
	We introduce the first challenging dataset UCIS4K for underwater camouflaged instance segmentation task, featuring a diverse array of images of camouflaged marine organisms. 
	Meanwhile,  we propose the UCIS-SAM model, which incorporates three key components: CBOM for underwater knowledge learning to eliminate color distortion in underwater scenes, FDTIM for camouflage knowledge learning to isolate misleading or deceptive information, and MFFAM for enhancing the aggregation of multi-level camouflaged features across different frequencies for more accurate segmentation.
	Extensive experiments validate the effectiveness of the UCIS4K dataset and demonstrate UCIS-SAM's superior segmentation accuracy and robust generalization capability.

	\bibliographystyle{IEEEtran}
	\bibliography{wang1abbreviation}

\end{document}